\documentclass[10pt,twocolumn,letterpaper]{article}

\usepackage{iccv}
\usepackage{times}
\usepackage{epsfig}
\usepackage{graphicx}
\usepackage{amsmath}
\usepackage{amssymb}
\usepackage[utf8]{inputenc} 
\usepackage[T1]{fontenc}    
\usepackage{url}            
\usepackage{booktabs}       
\usepackage{amsfonts}       
\usepackage{nicefrac}       
\usepackage{microtype}      
\usepackage{xcolor}         

\usepackage{caption}
\usepackage{booktabs}
\usepackage{multirow}

\usepackage{lipsum}
\usepackage{array}
\usepackage{float}
\usepackage{bm}
\usepackage{subfigure}
\usepackage{algorithm}  
\usepackage{algorithmic}
\newcommand*{\ours}{Skeleton-Graph }

\usepackage[pagebackref=true,breaklinks=true,letterpaper=true,colorlinks,bookmarks=false]{hyperref}

\iccvfinalcopy 

\def\iccvPaperID{12} 

\begin{document}

\title{Skeleton-Graph: Long-Term \\ 3D Motion Prediction From 2D Observations Using Deep Spatio-Temporal Graph CNNs}

\author{
	Abduallah Mohamed \ \ \ 
	Huancheng Chen \ \ \ 
	Zhangyang Wang \ \ \ 
	Christian Claudel\\
	\small The University of Texas At Austin \ \ \ \ \ \
}





\maketitle
\ificcvfinal\thispagestyle{empty}\fi

\begin{abstract}
Several applications such as autonomous driving, augmented reality and virtual reality require a precise prediction of the 3D human pose. Recently, a new problem was introduced in the field to predict the 3D human poses from observed 2D poses. We propose Skeleton-Graph, a deep spatio-temporal graph CNN model that predicts the future 3D skeleton poses in a single pass from the 2D ones. Unlike prior works, Skeleton-Graph focuses on modeling the interaction between the skeleton joints by exploiting their spatial configuration. This is being achieved by formulating the problem as a graph structure while learning a suitable graph adjacency kernel. By the design, Skeleton-Graph predicts the future 3D poses without divergence in the long-term, unlike prior works. We also introduce a new metric that measures the divergence of predictions in the long term. Our results show an FDE improvement of at least 27\% and an ADE of 4\% on both the GTA-IM and PROX datasets respectively in comparison with prior works. Also, we are 88\% and 93\% less divergence on the long-term motion prediction in comparison with prior works on both GTA-IM and PROX datasets. Code is available at \url{https://github.com/abduallahmohamed/Skeleton-Graph.git}.
\end{abstract}
\section{Introduction}
An accurate 3D pose prediction model is vital for several applications. In intersection management and autonomous vehicles, one can prevent an accident based on the 3D poses of pedestrians~\cite{gu2019efficient,ding2018vehicle,rodriguez20183d}. In Virtual and Augmented Reality (VR/ AR) the predictions of the 3D pose help in deepening the immersive experience~\cite{pohl2019neural}. Where in drones and autonomous driving, 3D pose prediction helps in accurate maneuver and motion planning through the environment~\cite{huang2018act,zhou2018human,huang2019learning,8695392}. The process of obtaining the 2D poses is quite inexpensive in comparison to the process of obtaining the 3D poses. The 2D process does not require special sensors such as depth sensors installed on the device. Thus it is cheaper to obtain the 2D poses from an ordinary vision sensor~\cite{bkak2016recent}. However, 2D poses are not suitable for the increasing trends of requirements in a new range of applications such as the ones mentioned before~\cite{sarafianos20163d,rogez2016mocap}. Thus it is tempting to develop models that directly obtain 3D poses from 2D ones, saving the need for expensive equipment and upscaling to the new trends. 

The recent work of~\cite{cao2020long} introduced a new long-term trajectory prediction problem that focuses on the concept of obtaining 3D poses from 2D ones. The introduced problem goal is to predict the future expensive 3D motion trajectories from the cheaply obtained 2D observed motion. In their work, a deep model called GPP-Net was introduced to address this problem. It is a three-stage deep model that uses several concepts such as Variational Auto Encoders (VAEs) and tailored stages for both paths and poses predictions. Beside GPP-Net, other 3D human motion estimation models such as TR~\cite{vaswani2017attention}, VP~\cite{pavllo20193d}, LTD~\cite{wei2019motion} were evaluated on this problem. 
\begin{figure}[t]

\begin{center}
\includegraphics[width=\columnwidth]{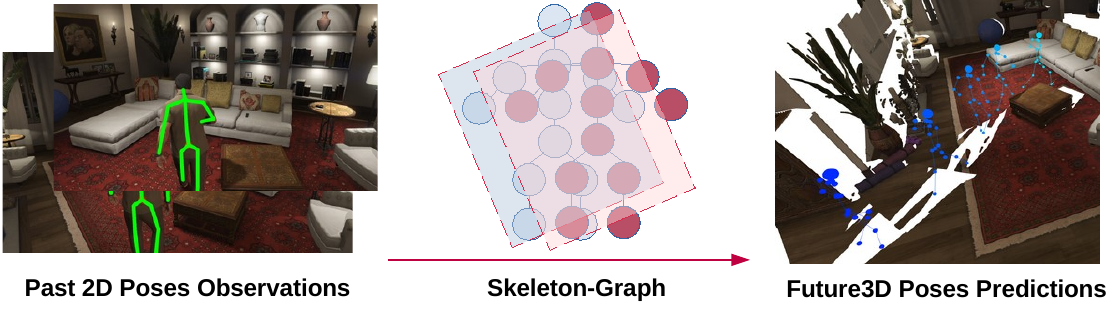}
\end{center}
   \caption{\ours given an input of 2D spatio-temporal graph of observed human skeleton poses. Then, in a single pass it predicts the next 3D poses.}
\label{gr:teaser}
\end{figure}

By examining the results and the architectures of these prior works, we found that three main design components were used separately in each prior work to enhance the results but not collectively in one work. The first component is the exploitation of the spatial configuration of the skeleton explicitly through the deep architecture itself~\cite{wei2019motion}. The skeleton spatial configuration includes useful information the leads to better predictions. The joints correlate with each other in terms of the angles and distances between them. This kind of information when introduced to a deep model it will constraints the model output not to produce random points that are far away from the ground truth. The second component is the usage of the vision signal~\cite{cao2020long}. The vision signal of the observed sequence contains information such as the objects in the environment and the geometry of the scene. This information helps in resolving the ambiguity in some prediction scenarios. For example, it is not valid to predict a skeleton standing on a table or a couch. The third and last component is to avoid the use of deep recurrent architectures. Recurrent architectures in long-term prediction problems tend to accumulate prediction errors from a step to the next step. We noticed this behavior in prior works~\cite{pavllo20193d,cao2020long,wei2019motion} that used these recurrent models.

Thus we introduce \ours in which we combine these three design components in one work overcoming the shortcoming of the prior works. First, to use the spatial configuration information of the skeleton we model the problem as a spatio-temporal graph end to end. We rely on the adjacency matrix to encode the relationship between the skeleton joints. Instead of a fixed adjacency matrix, we let our model learn it and analyze it in our ablation study. Secondly, we utilize the vision signal by fusing it into our model. Thus, we utilize the context information to enhance our results. We found out that the visual signal in some cases enhances short-term predictions. Lastly, to avoid divergence over the long-term we use a full CNN approach end to end without any recurrent behavior. This led to better long-term predictions in comparison with prior works. We introduce a new metric that measures the divergence over the prediction horizon to quantitatively judge this criterion of prediction stability.

This work is organized as follows, we start with the literature review of related 2D and 3D pose estimation methods, as well as deep graph models and trajectory prediction methods. Then, we follow up with the problem formulation and description of \ours method. Next, we discuss the problem of inconsistency in the prediction of the 3D skeleton poses and introduce a new objective that ensures the predicted 3D skeleton looks natural. We analyze the performance of our approach both quantitatively and qualitatively and discuss the prediction stability over the long term. 

\section{Related Work}
\textbf{2D pose estimation: }With the resurgence of neural nets, data-driven prediction paradigms have become more dominant. DeepPose~\cite{toshev2014deeppose} was the first major paper that applied deep learning to human pose estimation. In this approach, pose estimation is formulated as a CNN-based regression problem towards body joints. The work of~\cite{tompson2015efficient} generates heatmaps, describing the likelihood of the skeleton joints by running an image through multiple resolution banks in parallel to simultaneously capturing features at a variety of images scales.~\cite{newell2016stacked} Introduced a novel and intuitive architecture that consists of steps of pooling and upsampling layers to capture information at every scale. The previous approaches work well but were complicated in comparison with the next work. The work of~\cite{xiao2018simple} came up with a quite simple but efficient structure that consists of ResNet and few deconvolutional layers instead of the upsampling mechanism.

\textbf{3D pose estimation: }Recovering 3D pose from 2D RGB images is considered more difficult than 2D pose estimation, due to the larger 3D pose space and more ambiguities. The work of~\cite{li20143d} built a framework that consists of a joint points regression task and point detection task. They train a network that directly regresses 3D points from an image. Later,~\cite{chen20173d} explored a simple architecture that reasons through intermediate 2D pose estimations instead of directly estimating 3D pose from an image. Further~\cite{martinez2017simple} explored the sources of error in 3D pose estimation. They doubted the source of error is either from 2D to 3D mapping or from the improper visual analysis of the scene. They concluded that lifting the ground truth 2D joints locations to 3D space is not the source of error and it is a relatively straightforward task but the visual analysis is the main issue. However, since these methods completely ignore the image context, the predicted human motion may not be consistent with the scene. To perceive the scene context in the pose estimation task,~\cite{hassan2019resolving} exploits static 3D scene structure to better estimate human pose from monocular images considering environment constraints. A limitation of the current formulation is that it does not model scene occlusion. More recently, ~\cite{cao2020long} formulates a new task of long-term 3D human motion prediction with scene context in terms of 3D poses and develops a novel three-stage computational framework that utilizes scene context for goal-oriented motion prediction.

\textbf{Advances in deep graph models: }Recent advances in deep graph CNNs~\cite{kipf2016semi} lead to a jump in the domain of pose and trajectory predictions~\cite{mohamed2020social,huang2019stgat,wang2021graphtcn,cai2019exploiting,wang2021graphtcn} as graphs CNNs can exploit both the spatial configurations of the agents and their corresponding inner features. Several lines of work successfully classified human skeletons' actions by considering a spatio-temporal graph formulation of the skeleton~\cite{yan2018spatial,huang2020spatio}. The recent work of~\cite{li2020dynamic} proposed a multi-scale deep graph network to predict 3D human motion from 3D history.

\section{Problem Formulation and Technical Approach}
Given a pair of observations of $T$ time steps 2D human pose $P_\text{2D}$ and $T$ steps 2D images $I^\text{2D}$ of the scene the goal is to predict the next $\tilde{T}$ steps 3D human poses $P_\text{3D}$. The human skeleton have $J$ keypoints such that $P_\text{2D} \in \mathbb{R}^{J\times2}$ and $P_\text{3D} \in \mathbb{R}^{J\times3}$. Also, the pose prediction problem includes a path prediction problem within it. The path prediction problem is important because it is required to keep track of the 3D poses and to warp or re-normalize them when needed. The position of the path is the center of the torso.

\begin{figure*}[t]
\begin{center}
\includegraphics[width=\linewidth]{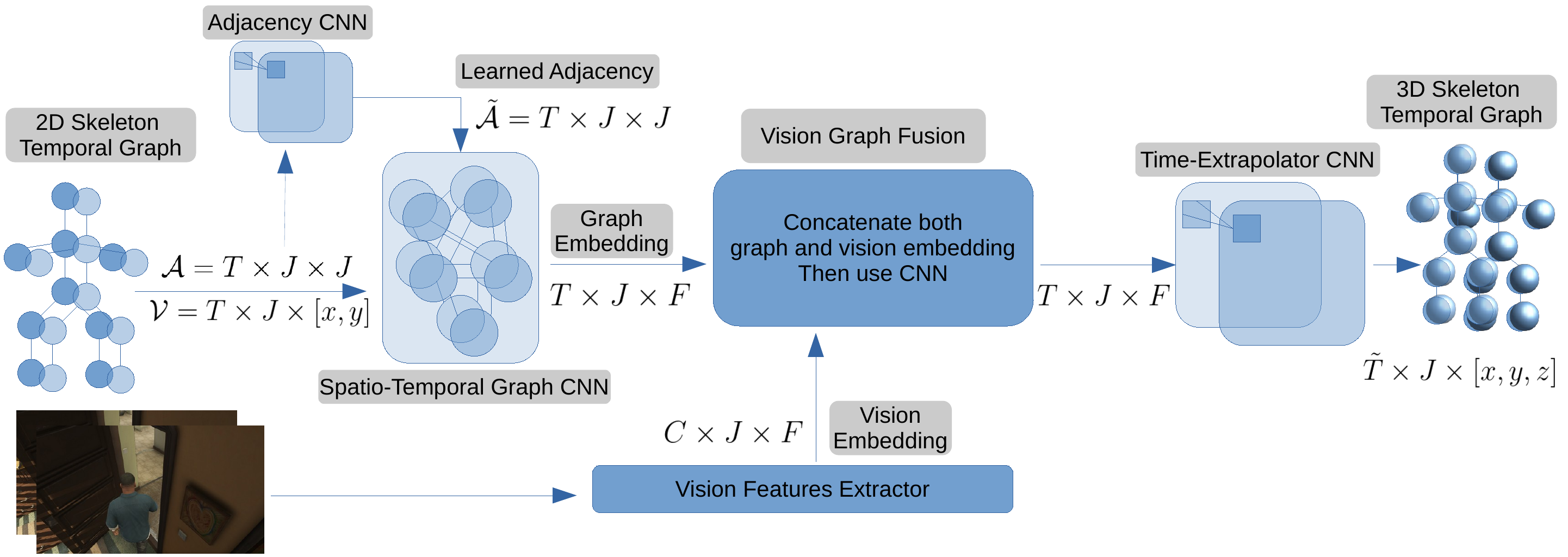}
\end{center}
   \caption{\ours model components. The model receives as an input a 2D skeleton temporal $T$ graph poses $\mathcal{V,A}$ and predicts the the next 3D skeleton temporal poses. The model can learn a suitable adjacency $\mathcal{\tilde{A}}$ matrix through the adjacency CNN. Also, it can use the observed visual signal in the form of a still image or a video. The time-extrapolator CNN is responsible for predicting the next $\tilde{T}$ temporal 3D poses, while the spatio-temporal graph CNN processes the 2D temporal graphs. $J, F, C, [x,y,z]$ are the number of the skeleton joints, the learned features dimensions, the vision features channels and the joint coordinates, respectively. }
\label{gr:model}
\end{figure*}
In what follows, we describe our approach to solve the problem at hand. We first start by modeling the problem as a spatio-temporal graph that represents the 2D motion of the human skeleton over time. Then we describe the \ours model wrapping with the loss function. 
\subsection{The formulation of the spatio-temporal graph}
We start modeling the problem as a spatio-temporal skeleton graph. On each observed time-step $t$, $\{t \in \mathbb{Z} \mid\, 0\leq t \leq T\}$ we define a graph $\mathcal{G}_t = \left(\mathcal{V}_t,\mathcal{E}_t\right)$. Where $\mathcal{V}_t$ is the graph vertices at time step $t$. Each vertex will hold the pose $P_\text{2D}^t$ information of the corresponding $j$, $\{j \in \mathbb{Z} \mid 0\leq j < J\}$ skeleton key-point. The $\mathcal{E}_t$ is the collection of graph edges. The adjacency matrix $\mathcal{A}_t$ defines the weight values of the graph edges. In our formulation we set the weights to be one, later we describe in our model how we learn proper values for entries of $\mathcal{A}_t$. 
Now, we can represent our input as a spatio-temporal graph $\mathcal{SG}_\text{input}= \{\mathcal{G}_t^\text{2D}\; |\; t\in \mathbb{Z},\, 0\leq t \leq T  \}$ and the output is $ \{\tilde{P}_\text{3D}^t\; |\; t\in \mathbb{Z},\, T < t \leq \tilde{T} \}$ which is the next 3D poses.

\subsection{\ours model}
From a top view, our model consists of four main components. The spatio-temporal graph CNN (SPGCNN) receives the spatio-temporal graph of the observed motion and generates a representation for it. The second component lies within the SPGCNN component. It is a CNNs that learns an adjacency matrix based on the temporal skeleton structure. The third component is the vision graph fusion. This component mixes both the graph representation from SPGCNN and the visual signal from the vision extractor. Lastly, the time-extrapolator CNN (TXCNN) receives this fused representation of both visual and graph and predicts the next 3D poses. In the upcoming section, we explain each component separately. These components can be seen in Figure~\ref{gr:model}. Our code is open-sourced and includes all the implementation details of the model. The supplementary materials include the fine implementation details of each component. 

\textbf{The spatio-temporal graph CNN (SPGCNN)} Our design of this layer processes the graph vertices data which has the shape of $T\times J \times [x,y]$ in two steps. The first step is a spatial step where it applies CNNs over the graph nodes weighting the CNN kernel by the values of the adjacency matrix. This spatial step takes into account the connection between the skeleton nodes by exploiting the adjacency matrix. This aligns with our design goal of using this information to enhance the results. Then, a temporal step is applied to the tensor from the spatial step. This step is a simple CNNs but acts on the time $T$ as a features channel. We refer to the work of \cite{yan2018spatial} regarding more details about the spatio-temporal graph CNNs. The output of this layer is a graph embedding that represents the observed 2D skeleton motion over time and has the shape of $T\times J \times F$, where $F$ is the learned features of each joint. 

\textbf{Learning a proper adjacency matrix}
The work of~\cite{mohamed2020social} proposed a kernel function to weigh the adjacency matrix that is tailored to their problem. Instead, in our approach we let the model discover the best weights of the adjacency matrix~\cite{wei2019motion}. The adjacency matrix $\mathcal{A}$ has the dimensions of $T \times J \times J$, one can imagine it as a 2D image with $T$ features channels. Thus, we use CNNs that take the temporal skeleton adjacency matrix as an input and learns a new one. Then, this learned adjacency $\mathcal{\Tilde{A}}$ is used within the previous component of our model. In our ablation study, we show the benefit of using this learned matrix. The learned adjacency matrix $\mathcal{\Tilde{A}}$ has the dimensions of $T \times J \times J$ but differs from the original one $\mathcal{A}$ in terms of the entries. This can be seen in Figure~\ref{gr:learnedadj}.

\textbf{Vision graph fusion}
\begin{figure}[t]
\begin{center}
\includegraphics[width=\columnwidth]{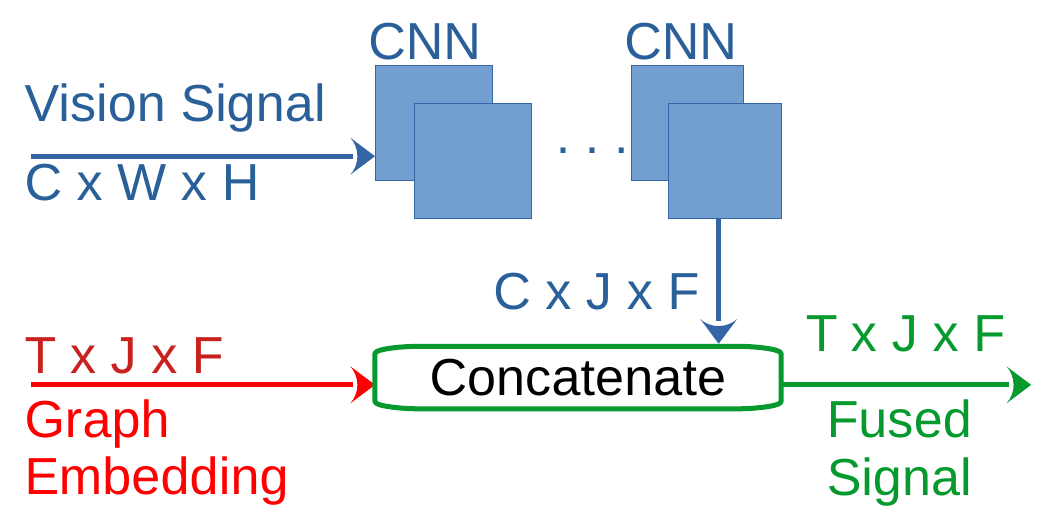}
\end{center}
   \caption{Vision graph fusion. $C, W, H$ are the image channels, width and height. }
\label{gr:fusion}
\end{figure}
As we observe 2D images $I^\text{2D}$, using them could be beneficial to our model. We have two options, the first as in~\cite{cao2020long} to use the last observed image only $I^\text{2D}_{T}$. The intuition behind using the last image is that it is the nearest observation of what is to be predicted next. We also tested the idea of using the full sequence of the observed images $I^\text{2D}_{0,..,T}$ to use more visual context. Yet, empirically as we will show in the experiments section using the last image improves the path prediction while the whole sequence of images improves the pose prediction on one of the datasets. The vision feature extractor is a CNNs that is designed to down-size the images in terms of the spatial aspects (Width and height) to match the graph embedding. Then, both the image embedding and the graph embedding are concatenated. This can be seen in Figure~\ref{gr:fusion}. This simple concatenation gives the model the ability to learn a fused representation leveraging both visual and graph contexts. The final representation has the dimensions of $T\times J \times F$. This representation is to be used by the next layer to predict the next $\tilde{T}$ 3D poses.

\textbf{The time-extrapolator CNN (TXCNN)}
This is the last step in \ours model. As we arrive at the embedding that represents the history of the 2D observations, we attempt to predict the next $\Tilde{T}$ 3D steps. As in \cite{mohamed2020social,bai2018empirical,malla2020social,wang2021graphtcn,zhao2020noticing}, using CNNs was proven to be better than recurrent models as it does not result in diverged predictions. The TXCNN receives an embedding of the history with the shape of $T \times J \times F$. The TXCNN treats the time as a features channel and through ordinary CNNs, it predicts the next $\Tilde{T}$ 3D poses $\tilde{P}$ steps. Simply, it extrapolates the time into the future moving it from $T$ observed steps to $\tilde{T}$ predicted steps. The final output of our model is $\tilde{T} \times J \times [x,y,z]$ which is the predicted 3D poses.
\section{The Skeleton Consistency Objective}
\begin{figure*}[t]
	\centering
	\includegraphics[width=\linewidth]{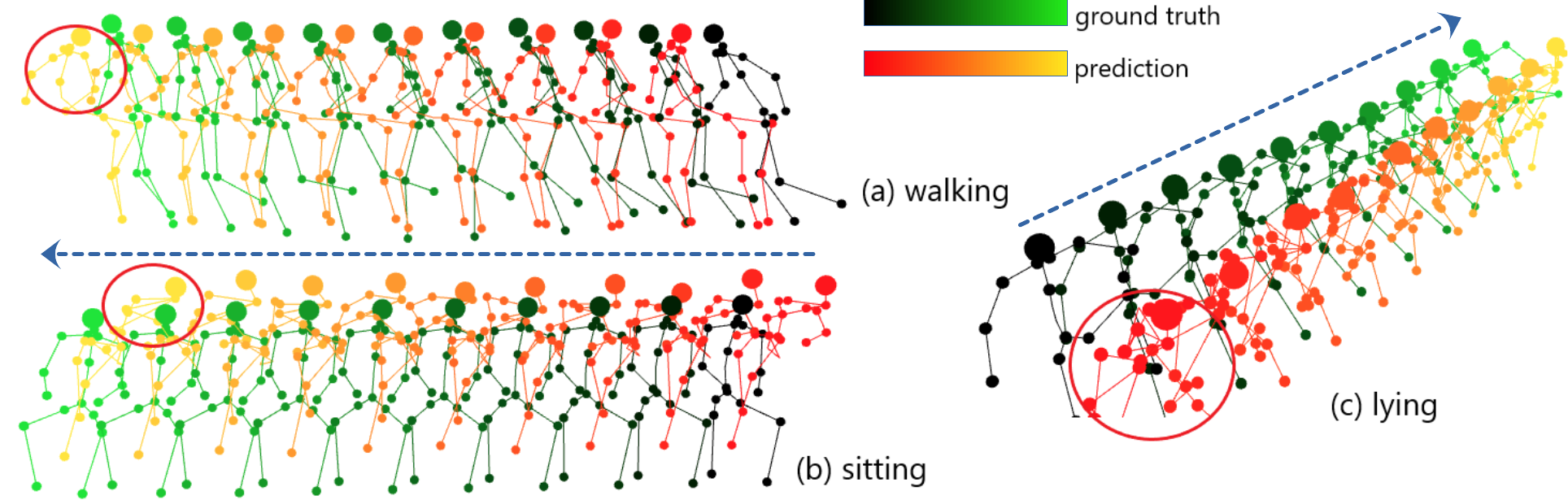}
	\caption{Different cases of deformation of the predicted 3D skeletons. Arrow indicates the time direction. The first two poses (a) and (b) show awkward joint angles between each joint. The case (c) one the right shows failure in both predicted angle and bone length.}
	\label{gr:weirdskeletons}
\end{figure*}

During our experiments, we noticed that we have accurate results exceeding the state-of-the-art but the predicted 3D skeletons do not look natural. For example, a skeleton might be setting but the distance between the neck and body is awkward. Examples of these abnormal skeletons can be seen in Figure~\ref{gr:weirdskeletons}. Due to this, we introduce a loss function that forces the predicted skeletons to be more consistent. We call this loss Skeleton Consistency Loss (SCL) which enforces the correct bone length and angles between the joints in the predicted 3D poses. The skeleton consistency concept was introduced in prior works~\cite{brau20163d, dabral2018learning, ramakrishna2012reconstructing, bkak2016recent,shi2020motionet} in which they focus on the skeleton reconstruction problem not the 3D pose prediction problem. For the bone length, the prior approaches and ours use $L_2$ norm to force the correct bone length. For the angles between the joints, some used the rotation angles but we preferred to use the cosine similarity because of its well-defined range. As mentioned before our consistency loss is composed of two parts. The first part is a cosine similarity between each skeleton joint $J$ in comparison with the predicted joints. It is defined as follows: 
\[
    \resizebox{0.49\textwidth}{!}{$\text{SCL}_{\text{cos}} = \frac{1}{\tilde{T}(J-1)}\sum^{t \in\tilde{T}}  \sum_{i = 1 }^{J-1} \left \|\mathcal{C} \left(P^t_i,P^t_{i+1}\right) - \mathcal{C} \left(\tilde{P}^t_i,\tilde{P}^t_{i+1}\right) \right\|_{1}$}
\]
Where $\mathcal{C}$ is the cosine similarity. The second part of the SCL is what enforces the reasonable bone length. It is defined as follows: \\ 
\[
\resizebox{0.49\textwidth}{!}{$\text{SCL}_{L_2} =\frac{1}{\frac{\tilde{T}J(J-1)}{2}}\sum^{t \in\tilde{T}} \sum^{j\in J} \sum_{i = j+1 }^{J}  \left\| \left\|P^t_i-P^t_j\right\|_2 - \left\|\tilde{P}^t_i-\tilde{P}^t_j\right\|_2 \right\|_{1}$}
\]
Thus, the SCL loss function  will be defined as: 
\begin{equation}
    \mathcal{L}_{\text{SCL}} = \lambda_1 \text{SCL}_{\text{cos}} +  \lambda_2 \text{SCL}_{L_2}
\end{equation}
Where $\lambda_1$ and $\lambda_2$ are weighting parameters. We found that setting both $\lambda_1=0.0005$ and $\lambda_2=0.1$ works the best.
 
Now, we can define the objective function that we train against:
\begin{equation}
    \mathcal{L_\text{\ours}} = \frac{1}{\tilde{T}J} \sum_{t =1 }^{\tilde{T}} \sum_{j =1 }^J \left \| \tilde{P}^t_j-P^t_j \right \|_2^2 + \mathcal{L}_{\text{SCL}} 
\end{equation}

\section{Experiments}
In this section, we describe the datasets used in the training, the training settings, and evaluation metrics. Then we compare with several prior models followed by an ablation study of our model both in quantitatively and qualitatively manners.
\subsection{Datasets description}
\noindent \textbf{GTA-IM:} GTA Indoor Motion Dataset~\cite{cao2020long} emphasizes on human-scene interactions. The motivation of this dataset is to fix the problem that real datasets of human-scene interaction has which is the noisy 3D human pose annotations and limited long-range human motion. The synthetic data of motions and interactions were collected from 3D video game \emph{Grand Theft Auto V} by controlling characters, cameras, and the physical system. The data set contains 50 human characters acting inside 10 different large indoor scenes. The dataset includes RGBD frames with $1920 \times 1080$ resolution, the corresponding ground-truth 3D human pose joints, human skeleton segmentation, and the camera parameters. We split 8 scenes for training and 2 scenes for evaluation following the settings of~\cite{cao2020long}. We also transfer the 3D path into the camera coordinate frame for both training and evaluation.
\noindent \textbf{PROX:} Proximal Relationships with Object eXclusion (PROX) is a new dataset captured using the Kinect-One sensor by~\cite{hassan2019resolving}. It contains 12 different 3D scenes with a total of 60 recorded scenarios. Each video is 30 FPS with camera parameters, calibration parameters, and human body segments. 3D skeleton points(25 joints) of the human pose are captured by Kinect-One sensor and 2D keypoints(25 joints) are captured by \href{https://github.com/CMU-Perceptual-Computing-Lab/openpose}{OpenPose}. In our experiment, we split PROX dataset with 52 training sequences and 8 sequences for testing following the settings of~\cite{cao2020long}. The usage of OpenPose to generate the ground truth makes it less accurate than the GTA-IM dataset. This is because the ground truth will inherit the errors of OpenPose. Also, the PROX dataset was mostly captured in a lab environment making it less diverse in terms of visual features. These flaws of the PROX dataset will impact our results as we will see in the ablation study section. A similar discussion was raised by the authors of~\cite{cao2020long}. Also, having results on the PROX dataset shows the robustness of our method in the case of noisy estimated 2D poses.

\subsection{Evaluation metrics}
The main metric for evaluating the 3D path and 3D pose prediction is the Mean Per Joint Position Error (MPJPE)~\cite{6682899}. For a frame $t \in \tilde{T}$ and a skeleton with $J$ joints , MPJPE is computed as:\begin{equation}E_{M P J P E}(t)=\frac{1}{J} \sum_{j=1}^{J}  \left \| \tilde{P}^t_j-P^t_j \right \|_2\end{equation}
The prior formulation is used to compute the 3D pose error. In case of the 3D path error, a specif joint is chosen such as the center of the skeleton torso. We also use two common metrics that can be found in the literature of pedestrian trajectory prediction~\cite{alahi2016social,gupta2018social,mohamed2020social} to evaluate our performance. First, the Mean Average Displacement Error(ADE) defined in equation~\ref{eq:ade} which judges the overall performance of both pose and path errors overall the predicted trajectory. Second, the Mean Final Displacement Error(FDE) which judges the performance at the final time step of the trajectory. The FDE is an indicator of the accumulation of the errors in the prediction, in other terms a low FDE means fewer errors were accumulated. The FDE defined in equation~\ref{eq:fde}. Lastly, we define a measure for the stability over the prediction horizon $\text{STB}_{\sigma}$ Equation~\ref{eq:spread}. It is the average error for both path and pose predictions. If the predictions deviate or accumulate errors over the long-term this metric increases and vice versa. Though this metric only discusses the divergence over the prediction horizon and it does not indicate the accuracy of the predictions, unlike the ADE and FDE metrics. 
\begin{equation}
\label{eq:ade}
\resizebox{0.5\textwidth}{!}{$\text{ADE}=\frac{1}{2}\left(\left[\frac{1}{\tilde{T}} \sum_{t=1}^{\tilde{T}} E_{M P J P E}(t) \right]_\text{pose} +\left[\frac{1}{\tilde{T}} \sum_{t=1}^{\tilde{T}} E_{M P J P E}(t) \right]_\text{path}\right)$}
\end{equation}

\begin{equation}
\label{eq:fde}
\resizebox{0.5\textwidth}{!}{$\text{FDE}=\frac{1}{2}\left(\left[ E_{M P J P E}(t=\tilde{T}) \right]_\text{pose} +\left[ E_{M P J P E}(t=\tilde{T}) \right]_\text{path}\right)$}
\end{equation}

\begin{equation}
\label{eq:spread}
\resizebox{0.5\textwidth}{!}{$\text{STB}_{\sigma}=\sqrt{\frac{1}{2}\left(\text{Var}\left\{ E_{M P J P E}(t)| t\in\tilde{T}\right\}_\text{pose} +\text{Var}\left\{ E_{M P J P E}(t)|t\in \tilde{T} \right\}_\text{path}\right)} $}
\end{equation}

\subsection{Training settings}
For all of our experiments, we use SGD optimizer. The initial learning rate for GTA-IM is 0.01 and 0.03 for the PROX dataset. The number of training epochs is 450 and we decrease the learning rate by a factor of 0.2 every 200 epochs. We use a batch size of 128 and 1 second of observation and 2 seconds for predictions following the settings of~\cite{cao2020long}. We used GTX-1080Ti for training on a 128 GB RAM machine. The need for a large RAM comes when the models are trained using the visual signal. Training each model took between 8 hours and 24 hours depending on the used dataset.
\subsection{Comparison with Prior Methods on GTA-IM \& PROX Datasets}
In this section, we perform quantitative evaluations of our method. The evaluation results of the 3D path and pose predictions on the GTA-IM dataset are shown in Table~\ref{tab:gta_all} while the PROX dataset results are in Table~\ref{tab:prox_all}. Overall, \ours outperforms the prior methods on several metrics. For the FDE we are 105 mm more accurate than GPP-Net~\cite{cao2020long} on the GTA-IM dataset and 110 mm more accurate than GPP-Net on the PROX dataset. This shows that we did not accumulate error over the long-term prediction, unlike prior methods. The ADE is slightly better than the previous state-of-the-art GPP-Net by ~10mm. This means, on average we have a more accurate 3D path and pose predictions. For the divergence in the long-term, the $\text{STB}_\alpha$ has a drastic drop in comparison with prior works. We are 88\% better on the GTA-IM dataset and 93\% better on the PROX dataset. This can be seen in the tables where our 0.5 seconds are close to the 2 seconds prediction in terms of MPJPE which means no divergence happening in the long-term, unlike prior works. We also notice that our model on the 0.5 seconds horizon in GTA-IM does not perform better than the prior methods. The same notice can be seen in the PROX dataset results. We highlight this as a trade-off between accuracy over short-term prediction versus the stability of prediction over the long-term due to not using recurrent approaches. As discussed in the introduction recurrent approaches tend to be accurate in the short term. For example, we notice that prior methods tend to be accurate over the short-term but diverge drastically over the long term, unlike our model. Overall, though our model behaves like this in the short term, the overall average performance is still better than the prior methods by at least 27\% and 4\% on both the FDE and ADE metrics, respectively.

\begin{table*}[ht]
\centering
\scriptsize
\begin{tabular}{lccccccccccc} 
\toprule
                                           & \multicolumn{4}{c}{3D path error (mm)}                    & \multicolumn{4}{c}{3D pose error (mm)} & \multicolumn{3}{c}{$\downarrow$}                                                  \\
Time step (second)                         & 0.5          & 1            & 1.5          & 2            & 0.5         & 1            & 1.5          & 2            & \textbf{FDE}  & \textbf{ADE}  &   $\textbf{STB}_{\sigma} $  \\ 
\midrule
TR~\cite{vaswani2017attention}                        & 277          & 352          & 457          & 603          & 291         & 374          & 489          & 641          & 622 &436  &147                     \\
TR~\cite{vaswani2017attention} + VP~\cite{pavllo20193d}  & 157          & 240          & 358          & 494          & 174         & 267          & 388          & 526          & 510 &326  &150                     \\
VP~\cite{pavllo20193d} + LTD~\cite{wei2019motion} & 124          & 194          & 276          & 367          & 121         & 180          & 249          & 330          & 349 &230   &98                     \\
GPP-Net~\cite{cao2020long}                                    & \textbf{104} & \textbf{163} & 219 & 297 & \textbf{91} & \textbf{158} & 237 & 328 & 313&200   &93           \\ 
\midrule
\ours (ours)                                      & 154          & \textbf{163}          & \textbf{172}          & \textbf{186}          & 198          & 209          & \textbf{217}          & \textbf{230}          & \textbf{208} &\textbf{192}     &\textbf{11}                  \\
\bottomrule
\end{tabular}
\caption{  {\bf Results of the GTA-IM dataset}. Results of 3D path and pose MPJPE error are reported in mm. The lower, the better.}
\label{tab:gta_all}

\end{table*}


\begin{table*}[ht]
\centering
\scriptsize
\begin{tabular}{lccccccccccc} 
\toprule
                       & \multicolumn{4}{c}{3D path error (mm)}                    & \multicolumn{4}{c}{3D pose error (mm)} & \multicolumn{3}{c}{$\downarrow$}                                                 \\
~Time step (second)    & 0.5          & 1            & 1.5          & 2            & 0.5          & 1            & 1.5          & 2            & \textbf{FDE}  & \textbf{ADE} &   $\textbf{STB}_{\sigma} $  \\ 
\midrule
TR~\cite{vaswani2017attention}                        & 487          & 583          & 682          & 783          & 512          & 603          & 698          & 801          & 792  &644    &126                  \\
TR~\cite{vaswani2017attention} + VP~\cite{pavllo20193d}                & 262          & 358          & 461          & 548          & 297          & 398          & 502          & 590          & 569   &427      &126               \\
VP~\cite{pavllo20193d} + LTD~\cite{wei2019motion}              & 194          & 263          & 332          & 394          & 216          & 274          & 335          & 394 & 394    &300   &82                 \\
GPP-Net~\cite{cao2020long}                 & \textbf{189} & \textbf{245} & 317 & 389 & \textbf{190} & \textbf{264} & 335 & 406          & 398  &292  &90           \\ 
\midrule
\ours (ours) & 264          & 269          & \textbf{272}          & \textbf{277}          & 281          & 287          & \textbf{291}          & \textbf{298}          & \textbf{288}  &\textbf{280}  &\textbf{6}                    \\
\bottomrule
\end{tabular}
\caption{  {\bf Results of the PROX dataset}. Results of 3D path and pose MPJPE error are reported in mm. The lower, the better. }
\label{tab:prox_all}
\end{table*}
\section{Ablation Study}
In this section, we extensively analyze the different behaviors of \ours model in both quantitative and qualitative ways. We start with a quantitative analysis of the different configurations of our model such as the adjacency learning and the skeleton consistency loss. We follow up with a visual analysis of these components and their effect on the results.

\subsection{Quantitative analysis of model components}
In this section we study the different components of \ours model. The components are: Learning adjacency CNN $+\mathcal{\tilde{A}}$, image embedding $+I^\text{2D}_{T}$, video embedding $+I^\text{2D}_{1..T}$ and the two different components of our skeleton consistency loss $\text{SCL}_{\text{cos}}$, $\text{SCL}_{L_2}$. Table~\ref{tab:model_config} shows an evaluation of these configuration on both GTA-IM and PROX datasets. \\
\begin{table*}[ht]
\centering
\tiny
\begin{tabular}{l|lllll|lllllllllll} 
\toprule
\multirow{2}{*}{Dataset} & \multicolumn{5}{c|}{\ours configuration} & \multicolumn{4}{c}{3D path error (mm)}                                                                                                    & \multicolumn{4}{c}{3D pose error (mm)}    &   \multicolumn{3}{c}{$\downarrow$}                                                                                                                                             \\
                         & $+\mathcal{\tilde{A}}$ & +$\text{SCL}_{\text{cos}}$ & +$\text{SCL}_{L_2}$ & $+I^\text{2D}_{T}$ & $+I^\text{2D}_{0..T}$         & \multicolumn{1}{c}{0.5}          & \multicolumn{1}{c}{1}            & \multicolumn{1}{c}{1.5}          & \multicolumn{1}{c}{2}            & \multicolumn{1}{c}{0.5}          & \multicolumn{1}{c}{1}            & \multicolumn{1}{c}{1.5}          & \multicolumn{1}{c}{2}            & \multicolumn{1}{c}{\textbf{FDE}  } &
                          \multicolumn{1}{c}{\textbf{ADE}  } &   $\textbf{STB}_{\sigma} $
                         \\ 
\cmidrule{1-17}
\multirow{7}{*}{GTA-IM}  &     &     &      &     &                 & 150                                & 165                                & 175                                & 191                                & 198                                & 211                                & 220                                & 234                                & \multicolumn{1}{c}{213} &\multicolumn{1}{c}{193}     &16                                         \\
                         & \checkmark   &     &      &     &                 & \multicolumn{1}{c}{159}          & \multicolumn{1}{c}{169}          & \multicolumn{1}{c}{178}          & \multicolumn{1}{c}{192}          & \multicolumn{1}{c}{196}          & \multicolumn{1}{c}{208}          & \multicolumn{1}{c}{216}          & \multicolumn{1}{c}{229}          & \multicolumn{1}{c}{211}    & \multicolumn{1}{c}{193}      &14                \\
                         &\checkmark  &\checkmark  &      &     &                 & \multicolumn{1}{c}{161}          & \multicolumn{1}{c}{169}          & \multicolumn{1}{c}{176}          & \multicolumn{1}{c}{187}          & \multicolumn{1}{c}{199}          & \multicolumn{1}{c}{209}          & \multicolumn{1}{c}{217}          & \multicolumn{1}{c}{230}          & \multicolumn{1}{c}{208}  & \multicolumn{1}{c}{193}    &10                    \\
                         &\checkmark  &     &\checkmark   &     &                 & \multicolumn{1}{c}{159}          & \multicolumn{1}{c}{169}          & \multicolumn{1}{c}{177}          & \multicolumn{1}{c}{191}          & \multicolumn{1}{c}{\textbf{191}}          & \multicolumn{1}{c}{204}          & \multicolumn{1}{c}{212}          & \multicolumn{1}{c}{225} & \multicolumn{1}{c}{208}    & \multicolumn{1}{c}{191}    &14                  \\
                         &\checkmark  &\checkmark  &\checkmark   &     &                 & \multicolumn{1}{c}{159} & \multicolumn{1}{c}{168} & \multicolumn{1}{c}{177} & \multicolumn{1}{c}{190} & \multicolumn{1}{c}{193} & \multicolumn{1}{c}{205} & \multicolumn{1}{c}{213} & \multicolumn{1}{c}{226}          & \multicolumn{1}{c}{208}   & \multicolumn{1}{c}{191}  & 11            \\
                         &\checkmark  &  \checkmark   &  \checkmark    &\checkmark  &                 & \multicolumn{1}{c}{\textbf{154}}                                & \multicolumn{1}{c}{\textbf{163}}                                & \multicolumn{1}{c}{\textbf{172}}                                & \multicolumn{1}{c}{\textbf{186}}                                & \multicolumn{1}{c}{198}                               & \multicolumn{1}{c}{209}                                & \multicolumn{1}{c}{217}                               & \multicolumn{1}{c}{230}                                & \multicolumn{1}{c}{208}    &  \multicolumn{1}{c}{191}   &11                                      \\
                         &\checkmark  &  \checkmark   & \checkmark     &     &\checkmark              & \multicolumn{1}{c}{165}          & \multicolumn{1}{c}{174}          & \multicolumn{1}{c}{181}          & \multicolumn{1}{c}{193}          & \multicolumn{1}{c}{190}          & \multicolumn{1}{c}{\textbf{202}}          & \multicolumn{1}{c}{\textbf{211}}          & \multicolumn{1}{c}{\textbf{224}}          & \multicolumn{1}{c}{208} &\multicolumn{1}{c}{192}&11                        \\ 
\hline
\multirow{7}{*}{PROX}    &     &     &      &     &                 & 340                                & 348                                & 353                                & 360                                & 369                                & 375                               & 379                                & 386                               & 373  &364    &7                                        \\
                         &\checkmark  &     &      &     &                 & 309                                & 314                              & 317                                & 323                                & 335                                & 339                                & 342                                & 347                                & 335    &328   &5                                       \\
                         &\checkmark  &\checkmark  &      &     &                 & 280                                & 287                                & 290                                & 296                                & 297                                & 303                                & 307                                & 314                                & 305&296&5                                              \\
                         &\checkmark  &     &\checkmark   &     &                 & \textbf{264}                                & \textbf{269}                                & \textbf{272}                                & \textbf{277}                                & \textbf{281}                                & \textbf{287}                                & \textbf{291}                                 & \textbf{298}                               & 288 &280  &6                                           \\
                         &\checkmark  &\checkmark  &\checkmark   &     &                 & 293                               & 299                                & 302                                & 308                                & 292                                & 298                                & 302                                & 308                                & 308   &300 &5                                           \\
                         &\checkmark &\checkmark &\checkmark   &\checkmark   &                    & \multicolumn{1}{c}{353}                                & \multicolumn{1}{c}{353}                                & \multicolumn{1}{c}{355}                               & \multicolumn{1}{c}{360}                                & \multicolumn{1}{c}{358}                               & \multicolumn{1}{c}{362}                               & \multicolumn{1}{c}{365}                                & \multicolumn{1}{c}{370}              &365 &359&3                                \\
                         &\checkmark  &  \checkmark   &   \checkmark   &     &\checkmark              & \multicolumn{1}{c}{283}                               & \multicolumn{1}{c}{290}                                & \multicolumn{1}{c}{294}                                & \multicolumn{1}{c}{301}                                & \multicolumn{1}{c}{308}                                & \multicolumn{1}{c}{314}                                & \multicolumn{1}{c}{319}                                & \multicolumn{1}{c}{325}                                & 313   &304&6                                          \\
\bottomrule
\end{tabular}
\caption{{\bf\ours ablation}. $+\mathcal{\tilde{A}}$ is the model with learned adjacency.  $\text{SCL}_{\text{cos}}$ and +$\text{SCL}_{L_2}$ indicates the usage of the SCL components. The usage of last image is $+I^\text{2D}_{T}$, $+I^\text{2D}_{1..T}$ indicates the usage of whole observed sequence. All results are in mm, the lower the better. The (0.5, 1, 1.5, 2) are time steps in seconds. Bolded numbers are the best in each column.}
\label{tab:model_config}
\end{table*}

\begin{figure*}[t]
\scriptsize
\begin{center}
\includegraphics[width=\linewidth]{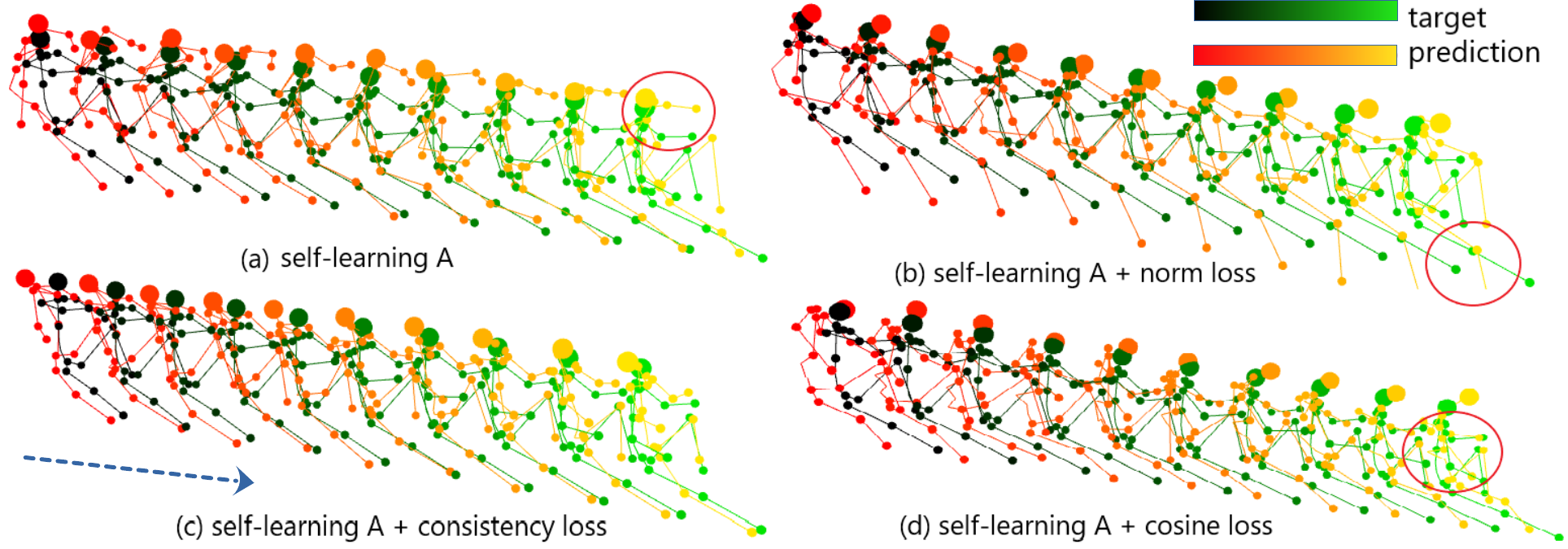}
\end{center}
   \caption{Effect of different configurations of \ours model on the 3D pose predictions. Arrow indicates the time direction. Top left:\ours model with learned adjacency $+\tilde{\mathcal{A}}$; Bottom left: \ours model $+\tilde{\mathcal{A}}$ + weighted $\mathcal{L}_{\mathrm{SCL}}$; Top right: \ours model  $+\tilde{\mathcal{A}}$ $+\mathrm{SCL}_{L2}$ only; Bottom right: \ours model  $+\tilde{\mathcal{A}}$ $+\mathrm{SCL}_{cos}$.}
\label{gr:qualti_ablation}
\end{figure*}
\begin{figure}[]
\small
\begin{center}
\includegraphics[]{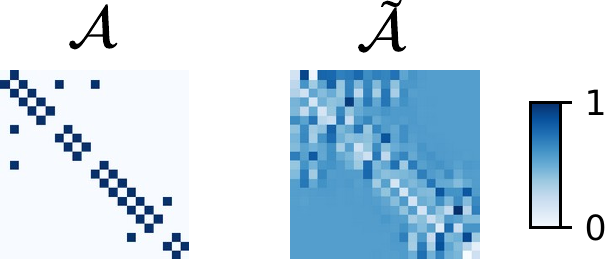}
\end{center}
   \caption{Heat map of the learned adjacency matrix $\tilde{\mathcal{A}}$ versus the original adjacency that represents the connection between the skeleton joints. }
\label{gr:learnedadj}
\end{figure}
\textbf{Plain \ours model: }
We directly train the raw skeleton-graph model without self-learning adjacent matrix, consistency constraints, and visual signal. This raw model alone outperforms the previous work on both the GTA-IM and PROX datasets on both FDE and ADE metrics. This validates that the graph approach we utilize in our model and the full CNN approach can remarkably decrease predictive MPJPE in the pose prediction task. We also notice that the results are divergence-free along the GTA-IM and PROX datasets with $\text{STB}_{\sigma} $ that is better than prior methods.

\textbf{\ours with learning adjacency $+\tilde{\mathcal{A}}$: }
The work of~\cite{mohamed2020social} suggested that the kernel function that governs the value of the adjacency matrix influences the results of the trajectory prediction significantly. Overall, the adjacency matrix is important in GCNNs because it governs the interaction between the graph nodes. Instead of searching for a handmade kernel function, we decided to let the model discovers the proper weights for the adjacency matrix. From Table~\ref{tab:model_config} we can see that the usage of the learned adjacency matrix $\text{STB}_{\sigma}$ improves the performance in comparison with the plain model. This indicates that the model discovered a better interaction between the graph nodes. Figure~\ref{gr:learnedadj} illustrates a heat map of the learned adjacency $\text{STB}_{\sigma}$. We notice an increase in the connections between the two legs joints, the same happens between the two hands joints. This emphasizes that the motion of humans has a strong pattern that is related to the coordination between both the hands and the legs. We also notice that the spine connections are almost the same as the original skeleton joints. This indicates that the spine does not contribute too much to the motion pattern.

\textbf{\ours +$\tilde{\mathcal{A}}$ + Skeleton Consistency Loss (SCL):}
Though the model with the learned adjacency matrix performs well, the model will generate many weird human poses without the consistency constraints. We show 3 failure cases in Figure~\ref{gr:weirdskeletons} where the angles and distances between joints are out of the normal range of a human body. 
We add cosine similarity$+\text{SCL}_{\cos}$ and joints distance$+\text{SCL}_{L_2}$ in the loss function and fine-tune hyper-parameters(optimal weights) to combine the three terms of loss: prediction-target loss, cosine similarity loss, and norm loss. We get improved performance by using these constraints as shown in Table~\ref{tab:model_config} on the GTA-IM dataset. On the other hand, in the PROX dataset, it seems only the$+\text{SCL}_{L_2}$ enhances the performance. This is because of the nature of the PROX dataset. The PROX dataset ground truth is machine-generated so it is not accurate when compared with the GTA-IM dataset which the ground truth is obtained from the game directly. This is behavior was seen before in the work of~\cite{cao2020long}. However, adding consistency loss does increase the training time in comparison with the plain model. Yet, it results in more accurate FDE and ADE results and more natural-looking skeletons as we will see in the qualitative study section.

\textbf{\ours +$\tilde{\mathcal{A}}$+SCL+ visual signal: }
We start with the GTA-IM dataset. The prior work of~\cite{cao2020long} shown that using the visual signal of the last observed frame enhances the performance. From Table~\ref{tab:model_config} we notice that using the last frame $+I^\text{2D}_{T}$ did enhance the short-term 3D path error in comparison with the previous components on the GTA-IM dataset. Yet, it did not enhance the short-term 3D pose error. When we used the full sequence of observed images $+I^\text{2D}_{0..T}$ on the GTA-IM dataset, the 3D pose error was the lowest among all the modes of configurations. This indicates that it helped predicting the 3D poses. Yet, for the 3D path, it had the highest error among all the configurations. This shows a trade-off between the path and poses objectives influenced by the presence of the visual signal. Overall, using the vision signal resulted in a performance that is similar to the usage of the consistency objective on the ADE and FDE metrics with a noticeable inner performance enhancement in the short term of both path and pose predictions. For the PROX dataset using the vision signal either the last image, $+I^\text{2D}_{T}$ or the full sequence $+I^\text{2D}_{1..T}$ resulted in a divergence of the results. This aligns with the findings of~\cite{cao2020long} over the PROX dataset. In addition, the PROX were captured in an empty lab environment, and thus it no as rich as the GTA-IM dataset. This made the dataset to be less diverse in terms of the background and the camera poses. So inherently the vision signal becomes less useful leading to ambiguity in the predictions as seen in Table~\ref{tab:model_config}.

\subsection{Qualitative study of the models' components}
To understand the effect of each configuration in our model, we visualize the predicted 3D human poses per each configuration as shown in Figure~\ref{gr:qualti_ablation}. Starting from the standard model with self-learning adjacent matrix alone $+\tilde{\mathcal{A}}$. The entire predicted pose looks natural in some areas and unnatural in other areas. The predicted right leg stretches out naturally and the left leg bends just like the target. However, the left hand in the pose is too close to the head and looks weird since there is no norm loss $\mathrm{SCL}_{L2}$ in the objective function. After we add norm loss(seen on the top right), the prediction has a reasonable left hand but the right leg joints penetrate the ground which is not possible in reality. Once the cosine loss being added $\mathrm{SCL}_{cos}$ to the objective, the penetration problem is gone but the distances between joints become unnatural especially in the first five frames(seen on the bottom right). When both $\mathrm{SCL}_{L2}$ and $\mathrm{SCL}_{cos}$ are used in training our model the resulted skeletons look very natural as shown in the bottom left in Figure~\ref{gr:qualti_ablation}. More qualitative cases are in the supplementary materials. 

\section{Conclusion}
We showed that the usage of graph CNNs with self-learning adjacency matrix and formulating the problem as a spatio-temporal graph is suitable for the problem of 3D skeleton motion predictions. We achieved state-of-the-art results on well-known benchmarks. The design of our model results in divergence-free predictions in the long term, unlike prior works. This was shown using the introduced $\text{STB}_\sigma$ metric. The deformation in prediction was solved using a skeleton consistency loss. The integration of the vision signal improved the results on the GTA-IM dataset. In the future, we would like to target the short-term prediction accuracy issue and investigate different methods for integrating the visual signals.

\textbf{Acknowledgement:} We thank the reviewers for their feedback. This research was supported by NSF (National Science Foundation) CPS No.1739964 and the CAMMSE UTC (US DOT).
\clearpage
{\small
\bibliographystyle{ieee_fullname}
\bibliography{refpaper}
}
\clearpage
\appendix

\section{Failure Cases}

Figure~\ref{tab:failure} illustrates several failure cases by our model. In case (1) we notice a skeleton sitting but our model was not able to capture the torso crouch to represent this mode. Case (2) the predicted skeleton trajectory overshoots, in other terms the model predicted too much of a momentum from the history of the observations. Case (3) and (4) are complex situations where the target skeleton changes from sitting to standing, we notice our model was not able to capture this trend. This is probably because of the lack of history that represents this trend in the 2D observations. Case (5) and (6) is a skeleton sitting or relaxing on an object, though our prediction are close, they look abnormal. This happens because in these two cases the joints angles vary a lot in which the predictions become harder. We believe in the future adding more dynamics oriented constraints can enhance these prediction errors.

\begin{table*}[ht]
\begin{tabular}{cccc}
(1) & \includegraphics[width=.4\linewidth]{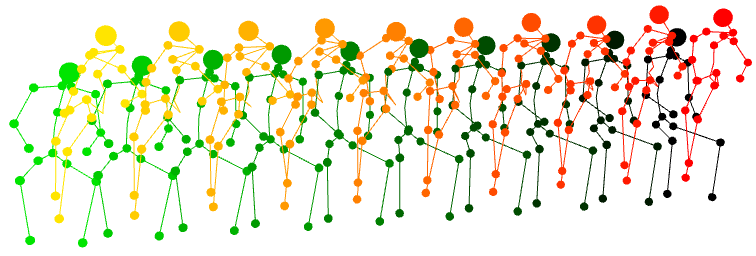} & (2)& \includegraphics[width=.4\linewidth]{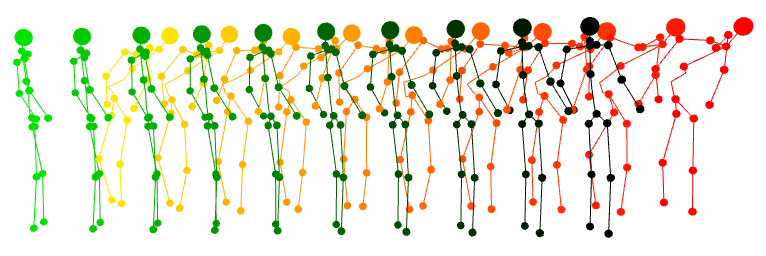}\\
(3) & \includegraphics[width=.4\linewidth,]{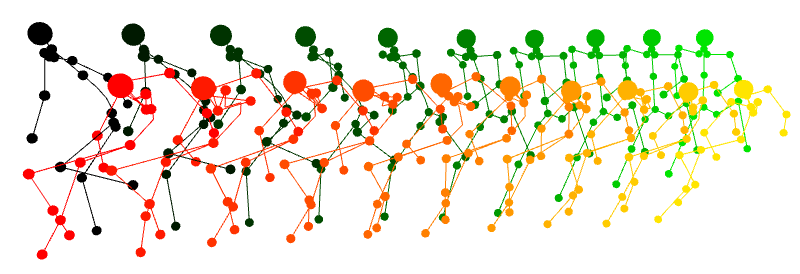} & (4)& \includegraphics[width=.4\linewidth]{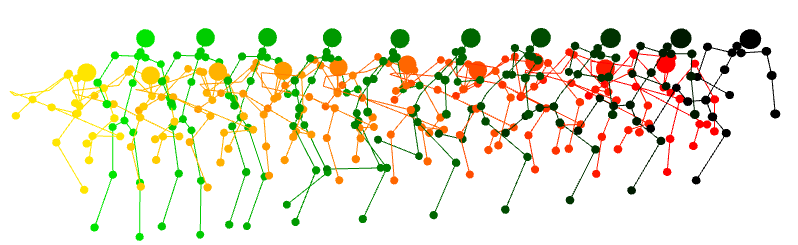} \\
(5) & \includegraphics[width=.4\linewidth]{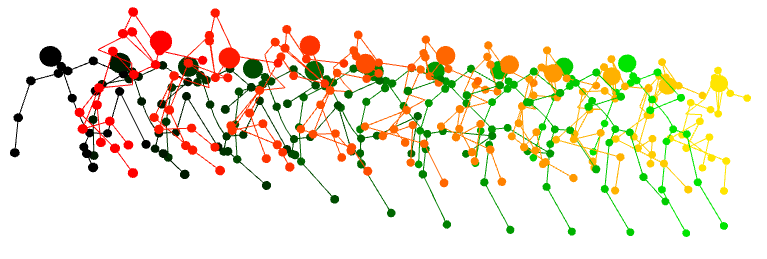} & (6) & \includegraphics[width=.4\linewidth]{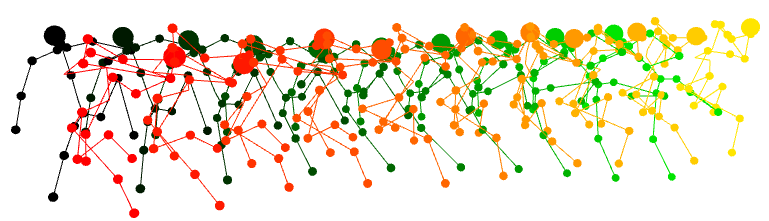} \\

\end{tabular}
\caption{Several failure cases predicted by \ours. The \textcolor{green}{green skeletons represents ground-truth} and \textcolor{red}{red skeleton represents prediction}. The deeper color the skeleton is, the earlier moment it is corresponding to. }
\label{tab:failure}
\end{table*}

\section{\ours Model Configuration}

\begin{table}[ht]
\centering
\begin{tabular}{l|c|ccccll} 
\toprule
Dataset                 & & 1       & 3       & 5       & 7       & 9       & 11                \\ 
\cline{1-1}\cline{3-8}
\multirow{4}{*}{GTA-IM} & 1                    & 320/362 & 267/290 & 211/247 & 221/250 & 213/237 & 230/245           \\
                        & 3                    & 262/296 & 252/278 & 230/251 & 198/246 & 212/239 & 210/225           \\
                        & 5                    & 272/298 & 273/277 & 234/257 & 206/234 & 203/241 & \textbf{202/231}  \\
                        & 7                    & 271/295 & 264/288 & 233/263 & 208/250 & 233/260 & 233/260           \\ 
\hline
\multirow{4}{*}{PROX}   & 1                    & 413/455       & 361/373       & 340/319       & 339/361      & 367/314      & 336/347                \\
                        & 3                    & 462/454       & 433/506       & 362/397      & 403/372      & 346/345      & 376/367              \\
                        & 5                    & 499/416       & 397/441       & 426/414      & 439/468      & 741/443      & \textbf{351/274}               \\
                        & 7                    & 582/679       & 455/420       & 400/408      & 420/426     & 409/484      & 371/351                \\
\bottomrule
\end{tabular}
\caption{ The effect of the choice of the number of TXCNN and SPGCNN on prediction results. The row is for the number of TXCNN layer while the column is for the number of SPGCNN layers. All readings are in mm, the lower the better.  The numbers are Path/Pose errors. The error is over 2 seconds prediction horizon.}
\label{tab:ablation}
\end{table}

We found out that the number of TXCNN and SPGCNN layers affects the model performance significantly. Table~\ref{tab:ablation} we show the effect with the number of TXCNN and SPGCNN on the model performance. We used residual connection while going deeper using this layers. First, we notice that going deeper with the number of SPGCNN results in a significant performance drop on both path and pose estimation. This is the same behavior noted in~\cite{li2019deepgcns}. Going deeper with the number of TXCNNs beyond 11 layers resulted in a drop of the performance. We notice that the performance for both 9 and 11 TXCNN layer are close to each other. Also, the result of the ablation is the same on both PROX and GTA-IM datasets. This means that our model is expected to behave in the same way on different datasets, a kind of agnostically to the dataset.

\section{Model Architecture}
Table~\ref{tab:arch} shows the inner details of our model. The structure of each component in terms of the parameters of the CNN layers, the usage of batch norm and the location of the activation functions are all shown in this table.

\begin{table*}[ht]
\centering
\begin{tabular}{l|l|l}
\toprule
Component                                  & Layer name / description & Layer strucutre                                                                                                                                                                                                                                       \\ 
\hline
\multirow{4}{*}{Spatio-Temporal Graph CNN} & Input                    & CNN(2,3,k=3,p=1)~                                                                                                                                                                                                                                     \\
                                           & Graph                    & GraphCNN(3,3)                                                                                                                                                                                                                                         \\
                                           & Learn Adjacency~         & CNN(3,3) BN,PReLU,CNN(2,3,k=3,p=1),BN~                                                                                                                                                                                                                \\
                                           & Temporal CNN~            & CNN(3,3) BN,PReLU,CNN(2,3,k=3,p=1),BN~                                                                                                                                                                                                                \\ 
\hline
Vision Features Extractor                  &                          & \begin{tabular}[c]{@{}l@{}}CNN(3,6,k=3,p=1), BN, PReLU,\\CNN(6,9,k=3,p=1,s=2), BN, PReLU,\\CNN(9,12,k=3,p=1,s=2), BN, PReLU,\\CNN(12,15,k=3,p=1,s=2), BN, PReLU,\\CNN(15,18,k=3,p=1,s=2), BN, PReLU,\\CNN(18,21,k=3,p=1,s=2), BN, PReLU\end{tabular}  \\ 
\hline
\multirow{2}{*}{Concatenate}               & \multirow{2}{*}{}        & Concatenate spatio-temporal graph CNN output~                                                                                                                                                                                                         \\
                                           &                          & with the vision features extractor output                                                                                                                                                                                                             \\ 
\hline
\multirow{3}{*}{Time Extrapolator CNN}     & Input                    & CNN(T+C,$\Tilde{T}$,3,p=1), BN,~PReLU                                                                                                                                                                                                                           \\
                                           & Middle                   & CNN($\Tilde{T}$,$\Tilde{T}$,k=3,p=1), BN,~PReLU + residual~                                                                                                                                                                                                               \\
                                           & Output                   & CNN($\Tilde{T}$,$\Tilde{T}$,k=3,p=1),                                                                                                                                                                                                                                     \\
\bottomrule
\end{tabular}
\caption{ \ours architecture description. CNN = Convolutional Neural Layer, k= kernel, p= padding, s= stride, BN= Batch Normalization, PReLU = Parametric ReLU activation function, T= observed time steps, C= vision signal features channels and $\Tilde{T}$= predicted time steps.}
\label{tab:arch}
\end{table*}
\end{document}